\documentclass[sigconf, nonacm]{acmart}

\usepackage{booktabs} % For formal tables
\usepackage{multirow}
\usepackage{icomma}
\usepackage{hyperref}
\usepackage{algorithm}
\usepackage{subcaption}
\usepackage{mathtools}
\usepackage[noend]{algpseudocode}

\makeatletter
\def\@copyrightspace{\relax}
\makeatother

\hypersetup{
	colorlinks,
	linkcolor={red!90!black},
	citecolor={green!80!black},
	urlcolor={blue!80!black}
}
\usepackage[capitalise,noabbrev]{cleveref}
\usepackage{appendix}

\setcopyright{none}
\usepackage{url}
\usepackage[utf8]{inputenc}
\usepackage{xcolor}
\usepackage[binary-units=true]{siunitx}
\usepackage{multicol}
\usepackage{rotating}
\usepackage[official]{eurosym}
\usepackage{color, colortbl}
\usepackage{listings}
\usepackage{balance}

\usepackage{tikz}
\usetikzlibrary{shapes,arrows,fit,calc}

\usepackage{syntax}

\usepackage[nolist]{acronym}

\newcommand{\ignore}[1]{}

\begin{document}
%\begin{appendices}
%  \input{appendices/travel_time.tex}
%  \input{appendices/congestion.tex}
%\end{appendices}
\sisetup{
	exponent-to-prefix = true        , %chktex 26
	round-mode         = figures     , %chktex 26
	round-precision    = 3           , %chktex 26
	scientific-notation = engineering
}

\begin{acronym}
	\acro{osm}[OSM]{OpenStreetMap}
	\acro{srtm}[SRTM]{Shuttle Radar Topography Mission}
	\acro{rfn}[RFN]{Relational Fusion Network}
	\acro{brr}[BRR]{Break Recovery Rate}
	\acro{sr}[SR]{Segmentation Rate}
	\acro{ch}[CH]{contraction hierarchy}
	\acrodefplural{ch}[CH]{contraction hierarchies}
	\acro{nos}[NOS]{non-optimal subpath}
	\acro{brp}[BRP]{Best Random Preference}
	\acro{rdp}[RDP]{Recovered Driving Preference}
	\acro{ttp}[TTP]{Travel Time Preference}
	%\acro{fpts}[FPTS]{Fastest Path Trajectory Segmentation}
	\acro{cts}[CTS]{Complete Trajectory Segmentation}
	\acro{opts}[OPTS]{Optimal Path Trajectory Segmentation}
	\acro{ppts}[PPTS]{Personalized Path Trajectory Segmentation}
	\acro{rrro}[RRRO]{Relative Recovered Route Overlap}
	\acro{rcrs}[RCRS]{Relative Cost Recovery Score}

\end{acronym}

\title{Scalable Unsupervised Multi-Criteria \\ Trajectory Segmentation and Driving Preference Mining}

\author{Florian Barth}
\affiliation{%
	\institution{Universit\"at Stuttgart, Germany}
}

\author{Stefan Funke}
\affiliation{%
	\institution{Universit\"at Stuttgart, Germany}
}

\author{Tobias Skovgaard Jepsen}
\affiliation{%
	\institution{Aalborg University, Denmark}
}

\author{Claudius Proissl}
\affiliation{%
	\institution{Universit\"at Stuttgart, Germany}
}

\begin{abstract}
	We present analysis techniques for large trajectory data sets that aim to provide a semantic understanding of trajectories reaching beyond them being point sequences in time and space.
	The presented techniques use a driving preference model w.r.t.\ road segment traversal costs, e.g., travel time and distance, to analyze and explain trajectories.

	In particular, we present trajectory mining techniques that can (a) find interesting points within a trajectory indicating, e.g., a via-point, and (b) recover the driving preferences of a driver based on their chosen trajectory.
	We evaluate our techniques on the tasks of via-point identification and personalized routing using a data set of more than 1 million vehicle trajectories collected throughout Denmark during a 3-year period.
	Our techniques can be implemented efficiently and are highly parallelizable, allowing them to scale to millions or billions of trajectories.
\end{abstract}

\maketitle

\section{Introduction}\label{sec:introduction}
The ubiquity of mobile devices with position tracking capabilities via GPS or localization using WiFi and mobile networks continuously generate vast streams of location data.
Such data may be used in a variety of ways.
Mobile networks providers and many companies, such as Google or Apple, use the location data of their customers to improve their services, e.g., by monitoring of traffic flow or detection of special events.
Location data sharing platforms such as Strava, GPSies, and \ac{osm} allow their users to share their location data with their community.
In all of these cases, location measurements are considered collectively as sequences, each reflecting the movement of a person or a vehicle.
Such sequences can be map-matched to paths in an underlying transportation network---in our case a \emph{road network}---using appropriate methods~\cite{Zheng2015}. We refer to such map-matched sequences as \emph{trajectories} throughout the paper.

A common assumption is that most of the time, users travel on `optimal' routes towards a (possibly intermediate) destination, where optimality is understood as the shortest path w.r.t.\  suitable scalar \emph{traversal costs} of each road segment in the underlying road network. For instance, route planners and navigation systems often use travel times as traversal costs.
However, in practice, drivers seldom travel on such `optimal' routes due to complex traversal costs, e.g., time-dependent and uncertain travel times~\cite{pedersen2020fast}, a (possibly unknown) combination of several traversal costs~\cite{DGG15}, or due to changing intentions/destinations during a trip.
We therefore investigate analysis techniques that do not rely on a fixed criterion but are capable of identifying a suitable combination of given criteria.

The high-level goal of this paper is to develop trajectory mining techniques to enable a better understanding of the semantics of trajectory data.
Concretely, we focus on the tasks of trajectory segmentation and driving preference mining.

\begin{figure}
	\includegraphics[width=0.95\linewidth, trim={0 3.925cm 0 3.925cm},clip]{%
		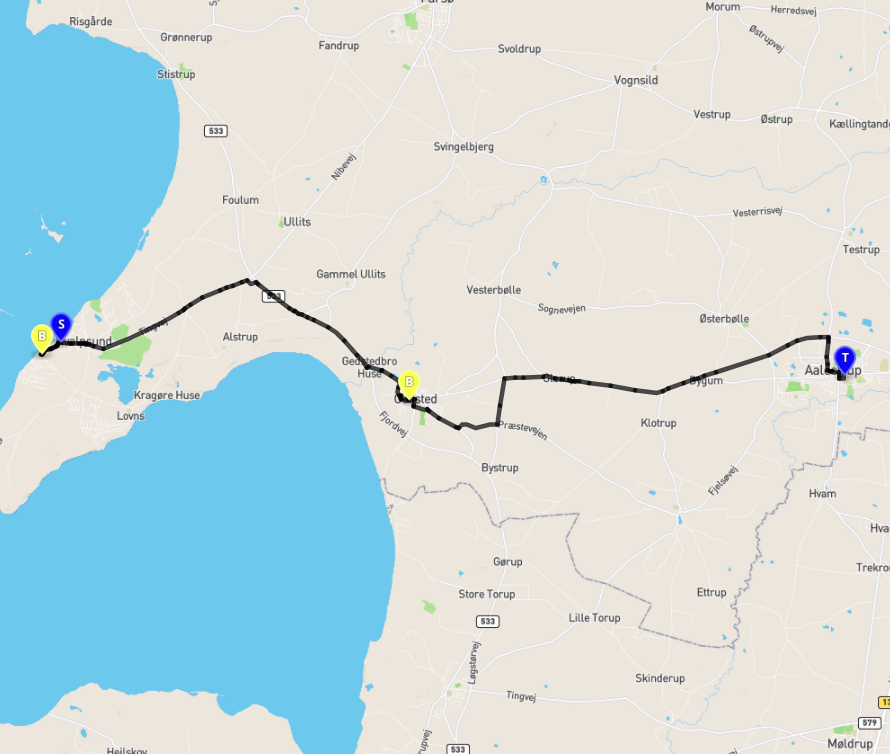}
	\caption{An example of a trajectory going from S to T with two intermediate stops that are labeled B.\label{fig:compelling-example}}
\end{figure}

\subsubsection*{Trajectory Segmentation}
A trajectory is often not just the manifestation of someone going from location S to location T following an optimal route w.r.t.\ some criteria, but rather determined by a sequence of activities/intentions.
For instance, \cref{fig:compelling-example} shows a trajectory from S to T with two intermediate stops labeled B.
The driver starts at location S but rather than taking the fastest routes, decides to drive southwest and makes a stop.
Then, the driver backtracks and takes the fastest route from S to T but decides to make a stop on the way.
In this paper, we present a trajectory segmentation approach that can identify intermediate stops or other points of interest in a trajectory and divide it into subtrajectories accordingly.

In contrast to previous work on trajectory segmentation~\cite{buchin2011segmenting,alewijnse2014framework,soares2015grasp,junior2018semi,etemad2019trajectory}, our approach solely relies on traversal costs and the structure of the road network. No additional information such as time stamps are required. Thus, compression techniques for efficient trajectory storage~\cite{krogh2016efficient,song2014press} are applicable.
However, despite not utilizing temporal information, our experiments show that our trajectory segmentation approach can recover such information through a structural analysis of the trajectory.
In addition, our trajectory segmentation approach uses a driving preference model to divide trajectories into subtrajectories.
To the best of our knowledge, this is the first trajectory segmentation approach to do so.

\subsubsection*{Driving Preference Mining}
Given a trajectory, it is an interesting question which criteria the driver likely tried to optimize.
We refer to this combination as a driving preference.
We show how a known algorithm for driving preference mining~\cite{FLS16} can be made sufficiently robust to recover preferences from vehicle trajectories.
We compare the modified algorithm with two benchmark functions and address the question, how well the recovered preferences describe the trajectories.

\subsection*{Related Work}
In this paper, we consider two main applications: trajectory segmentation and driving preference mining.

Previous work on trajectory segmentation can broadly be categorized as supervised~\cite{buchin2011segmenting,alewijnse2014framework} and unsupervised~\cite{soares2015grasp,junior2018semi,etemad2019trajectory}.
Supervised approaches require predetermined criteria.
In contrast, unsupervised approaches find the combination of available criteria that best explain the driving behavior in the input trajectory.
Our proposed trajectory segmentation approach is unsupervised. Thus, the required prior assumptions on drivers' behavior are reduced to a minimum.
In addition, our approach requires no spatio-temporal information about the trajectories, unlike existing trajectory segmentation approaches~\cite{buchin2011segmenting,alewijnse2014framework,soares2015grasp,junior2018semi,etemad2019trajectory}.

Driving preference mining---also referred to as driving preference learning ---has been studied previously as well.

% TODO: Related work to be integrated.
%Recovering the combination of two costs that make a trajectory optimal was considered in~\cite{Balteanu13}.
%In~\cite{DGG15}, a parameter learning approach is presented based on stochastic coordinate descent. They try to maximize trajectory overlap.
%A driving preference model based on Gaussian Mixture Models is described in~\cite{dai2015personalized}.
A popular approach of modeling driving preferences is to consider them as random variables. For instance, preference mining methods based on Gaussian Mixture Models are presented in~\cite{YangPersonalizedRouting} and in~\cite{dai2015personalized}. \citet{Balteanu13} present a probabilistic method that compares the input trajectory with pareto-optimal trajectory sets. \citet{FAVOUR} show how mining driving preferences can be accomplished using Bayesian learning strategies.

Our approach is non-probabilistic and considers the driving preference as the solution of an optimization problem. \citet{DGG15} present a similar technique, where the driving preference is chosen such that the overlap of the given and the computed trajectory is maximal. The main difference to our approach is that Delling et al.\ focus on geographical similarities of trajectories while we define similarity in terms of their traversal costs.

The work of \citet{FLS16} is the most closely related work both w.r.t.\ to trajectory segmentation and driving preference mining.
They present a method which decides whether there exists a conic combination of the traversal costs such that a given trajectory is optimal for this weighting---or \emph{preference}---of the traversal costs, and outputs the preference.
Their method can recover the driving preferences of synthetic trajectories where such a conic combination is guaranteed to exist.
However, this is not guaranteed for real world trajectories due to changes in driver preferences within the trajectory or inaccuracy in the traversal costs.
We extend their method to obtain robust preferences in case a trajectory is not optimal for any preference.
In addition, we use their method in our trajectory segmentation approach to determine the start and end of a subtrajectory.

% Toward personalized, context-aware routing.
% Deducing individual driving preferences for user-aware (may be cited already)
% Personalized and situation-aware multimodal route recommendations: The favour algorithm.

\subsection{Our Contribution}
%For our concrete data set \TODO{fill in some characteristics of the data} we are able to show that on average a single trajectory decomposes into only XXXXX 'optimal' subpaths when only travel time is considered. The decompositions often

We develop techniques to `explain' route choices made by drivers based on several natural criteria and evaluate them on a large data set of real-world trajectories.
In particular, we propose a simple method for unsupervised trajectory segmentation that is able to approximate locations where drivers change their intentions/destinations along their trajectories. In contrast to previous work, our approach does not require spatio-temporal information about the trajectories, which considerably reduces the storage requirements.

Additionally, we present a simple, yet effective modification of a driving preference mining technique that allows to estimate drivers' preference that is more robust with respect to noise or sporadic `suboptimal' routing decisions.

Both our approaches are built on the algorithm presented in~\cite{FLS16} which deduce driving preferences from trajectories.
As a result, driving preferences are integral to both our trajectory segmentation and driving preference mining techniques, demonstrating the close relationship between these two applications.
To the best of our knowledge, we are the first to show this relationship.

\section{Preliminaries}\label{sec:preliminaries}

\subsection{Data Set}\label{sec:data-set}

\subsubsection{Road Network Data}
We use a directed graph representation of the Danish road network~\cite{rfn-long} $G = (V, E)$ that has been derived from data provided by the Danish Business Authority and the \ac{osm} project.
In this graph representation, $V$ is a set of nodes, each of which represents an intersection or the end of a road, and $E$ is a set of edges, each of which represents a directed road segment.
The graph representation of the Danish road network contains the most important roads and has a total of $583,816$ intersections and $1,291,171$ road segments.
In addition, each road segment has attributes describing their length and type (e.g., motorway) and each intersection has attributes that indicate whether they are in a city area, a rural area, or a summer cottage area.
The data is further augmented with a total of $163,044$ speed limits combined from \ac{osm} data and speed limits provided by Aalborg Municipality and Copenhagen Municipality~\cite{rne-workshop}.

\subsubsection{Trajectory Data}
We use a set of $1,306,392$ vehicle trajectories from Denmark collected between January 1st 2012 and December 31st 2014~\cite{trajectory-data}.
The trajectories have been map-matched to the graph representation of the Danish road network s.t.\ each trajectory is a sequence of traversed road segments $T = (e_1, \dots, e_n)$ where $e_i \in E$ for $1 \leq i \leq n$.
In addition, each segment is associated with a time stamp and a recorded driving speed whenever the GPS data is sufficiently accurate.
In this data set, a trajectory ends after its GPS position has not changed more than $20$ meters within three minutes.
See~\cite{trajectory-data} for more details.

\paragraph{Trajectory Stitching}\label{sec:stitched-trajectories}
A vehicle trajectory in the trajectory data set ends when the vehicle has not moved more than $20$ meters within three minutes.
However, in practice, a driver may choose a trajectory with several intermediate stops, for instance when visiting multiple supermarkets to go grocery shopping.
We are interested in examining such trajectories.
We therefore stitch temporally consecutive trajectories from the same vehicle together if there is less than $30$ minutes difference between the end of the current trajectory to the start of the next.
Each stitch thus indicates the end of a $3$ to $33$ minutes break in movement. We call these stitches \emph{break points} that mark a temporal gap in the trajectory.

In many cases temporally consecutive trajectories are not connected due to imprecision or lack of GPS data.
In such cases, we compute the shortest route from the destination of the current trajectory to the start of the next.
If the shortest route is shorter than $200$ meters or consists of at most one road segment, we stitch the trajectories.
We continue attempting to join the stitched to the next trajectory until the next trajectory does not meet the stitching criteria.
See \cref{sec:trajectory-stitching} for further details.

From the original $1,306,392$ trajectories we obtain $260,190$ combined trajectories.
Of these trajectories, $190,199$ trajectories are stitched and contain break points.

\subsection{Routing Cost Types}\label{sec:metrics}
From the data sets described in \cref{sec:data-set}, we derive a number of criteria that are a measure of the expected cost of taking a route.
In our experiments, we use the following four cost types: travel time, congestion, crowdedness, and number of intersections. We normalize the average value of each cost type to one.

\paragraph{Travel Time}
Each road segment is associated with a fixed value that represents the time it takes to traverse the road segment.
To derive travel time, we combine historical traversal data from the trajectory data set with travel time estimates from a pre-trained machine learning model~\cite{rfn-long}.
See \cref{sec:travel-time-derivation} for further details.

\paragraph{Congestion}
We derive the congestion level on a particular road segment based on how close to the speed limit people tend to drive.
The closer to the speed limit, the less congestion.
Many road segments do not have a speed limit in our speed limit data set.
In such cases, we use a simple \ac{osm} routing heuristic, see \cref{sec:congestion-derivation}.

\paragraph{Crowdedness}
This criterion measures how `crowded' the surroundings along a vehicle trajectory are.
We derive a crowdedness value for each road segments from the number of nearby road segments and points of interest \ac{osm} nodes.
Further details can be found in \cref{sec:crowdedness-derivation}.

\paragraph{Number of Intersections}
The number of intersections visited in a trajectory, excluding the source intersection.

\subsection{Personalized Routing}
The notion of shortest path requires some distance or cost measure to compare paths (i.e., routes) in a road network.
To this end, we use the notion of personalized routing from~\cite{FS15} which takes into account both multiple traversal cost types and driving preferences.

\subsubsection{Personalized Cost}
The personalized cost of a road segment $e$ combines a $d$-dimensional \emph{cost vector} $c(e) = (c_1(e), \dots, c_d(e))$ and a $d$-dimensional \emph{preference vector} $\alpha=(\alpha_1, \alpha_2, \dots, \alpha_d)$ where $\alpha_i \geq 0$ and $\sum \alpha_i=1$.
Each cost $c_i(e)$ for $1 \leq i \leq d$ is a measure of the cost to traverse road segment $e$ and each preference $\alpha_i$ represent the drivers preference w.r.t.\ to minimizing cost $c_i(e)$.
For instance, $c_1(e)$ and $c_2(e)$ may be the travel time and the number of intersections of the road segment $e$.
Assuming equal scale of the costs, a preference vector of $(0.7, 0.3)$ indicates that a driver values travel time more than the number of intersections.

The personalized cost of a route $\pi = (e_1, e_2, \dots, e_k)$ in a road network is $p(\pi \mid \alpha) = \sum_{i=1}^k p(e_i \mid \alpha)$ where $p(e_i \mid \alpha)$ is the personalized cost of road segment $e_i$ and $\alpha$ represent the driving preferences of a particular driver.
Given driving preferences $\alpha$, the personalized cost of a road segment $e$ is $p(e \mid \alpha) = \sum_{i=1}^d \alpha_i c(e)$.
We call a path $\pi$ from $s$ to $t$ \emph{personalized path} if $\forall \pi' \in \Pi: p(\pi \mid \alpha) \leq p(\pi' \mid \alpha)$, where $\Pi$ is the set of all paths from $s$ to $t$.

\subsubsection{Deducing Routing Preferences}
Given a trajectory $T$ in the personalized route setting a natural question to ask is if it is a personalized path for some preference $\alpha$.
The trajectory $T= v_{0}v_{1}\ldots{} v_{k-1}$ is a personalized path if a solution exists to the following LP with variables $\alpha=(\alpha_1, \alpha_2, \dots, \alpha_d)$ representing driving preferences~\cite{FLS16}.
\begin{equation}\label{eq:linear-program}
	\begin{aligned}
		\textsc{Minimize} \qquad \ \ \, & 1                                                                        \\
		\textsc{subject to} \qquad      & \forall \pi \in \Pi \colon  p(T \mid \alpha)-p(\pi \mid \alpha)   \leq 0 \\
		                                & \sum_{i=1}^d \alpha_i =1                                                 \\
		                                & \forall i \in \{1, \dots, d\}\colon \alpha_i \geq 0
	\end{aligned}
\end{equation}
The LP contains a constraint for \emph{each} possible route $\pi \in \Pi$ from $v_0$ to $v_{k-1}$.
These constraints require the personalized costs of $T$ for the solution, the preference vector $\alpha$, to be lower than those of every other path $\pi$.
Note that no objective function is used and $T$ is therefore a personalized path if any preference vector $\alpha$ exists that satisfies all the constraints.

\paragraph{The Dijkstra Oracle}\label{sec:lp-path-oracle}
The solution to the LP in \cref{eq:linear-program} is a preference vector $\alpha$ for which no path from $v_0$ to $v_{k-1}$ has lower personalized cost than the trajectory $T$.
Writing down the complete LP for \emph{all possible} paths from $v_0$ to $v_{k-1}$ is infeasible.
Fortunately, it suffices to add the constraints one by one via a so-called separation oracle~\citep{FLS16}.

In brief, the LP is first solved using only the last two constraints in \cref{eq:linear-program}.\@
This results in some initial preference vector $\alpha$.
Then, $\alpha$ is verified to satisfy all the constraints in~\cref{eq:linear-program} by simply running a Dijkstra from $v_0$ to $v_{k-1}$ with preference vector $\alpha$.
If so, a solution to the linear program in \cref{eq:linear-program} has been found and no further processing is required. If not, a violating constraint is discovered and added to the LP.\@

While this method finds a preference method $\alpha$ for $T$, if any exists, real world trajectories are often not personalized paths.

\subsection{Trajectory Segmentation}\label{sec:segmentation-problem}
In this section, we discuss the definition of the trajectory segmentation problem and a general algorithmic framework for it.
\subsubsection{Trajectory Segmentation Problem}
The segmentation of a trajectory $T = v_0v_1\dots v_{k-1}$ is a sequence of $B$ trajectory segments $S_1=v_0 v_1 \dots v_{b_1}$, $S_2=v_{b_1} v_{b_1+1} \dots v_{b_2}$, $S_3=v_{b_2} v_{b_2+1} \dots v_{b_3}$, up to  $S_B=v_{b_{B-1}} v_{b_{B-1}+1} \dots v_{b_{B}}$.
We refer to the common node of two consecutive trajectory segments, e.g., $S_1$ and $S_2$, as a \emph{segmentation point}. For instance, $v_{b_1}$ is a segmentation point because it is at the end of $S_1$ and the start of $S_2$.
\citet{buchin2011segmenting}~define the segmentation problem as finding a (minimal) number of segments for a trajectory such that each segment fulfills a criterion.
They provide a general algorithmic framework for arbitrary segmentation criteria.

\subsubsection{Trajectory Segmentation Framework}\label{sec:segmentation-framework}
In the framework of \citet{buchin2011segmenting}, one has to provide a test procedure which verifies if a given segment meets the desired criterion.
This test procedure is then used to repeatedly, greedily find the longest prefix that meets the criterion.
The authors prove that this approach leads to an optimal, i.e., minimal, segmentation for \emph{monotone} criteria in $O(T(k) \log k)$ time if the test procedure takes $T(m)$ time for a segment of length $m$.
They define monotonicity for a criterion as follows.
If any segment $S \subseteq T$ satisfies the criterion, then any segment $S' \subseteq S$ also satisfies the criterion.
Even though~\citet{buchin2011segmenting} focus on self-similarity criteria for the segments, this definition adapts well to the optimality criterion we introduce in~\cref{sec:personalized-path-criterion}.

\section{Multi-Criteria Trajectory Segmentation}\label{sec:trajectory-segmentation}
A driver may have several via-points on a trip.
Sometimes, these via-points are linked to a point-of-interest such as gas station, but that is not necessarily the case.
The trajectory segmentation approach we present in this section is designed to find all interesting via-points along a trajectory.

In brief, our approach assumes that drivers choose personalized paths between their via-points.
Any deviation from their personalized path along a trajectory that goes from $s$ to $t$ indicates some interesting point $p$ in the trajectory.
The point of deviation $p$ is marked as the end of the first trajectory segment and the beginning of the next.
This process is repeated on the remaining subtrajectory going from $p$ to $t$ and so forth.

\subsection{The Personalized Path Criterion}\label{sec:personalized-path-criterion}
For trajectory segmentation in the framework of~\cite{buchin2011segmenting}, described in~\cref{sec:segmentation-framework}, we propose a type of criteria that uses only the underlying graph and does not need any predetermined parameters.
The \emph{optimal path criterion} requires each trajectory segment $S$ of a trajectory $T$ to be an optimal path according to the traversal costs in the underlying graph.
This criterion is monotone as defined in~\cref{sec:segmentation-problem} because it requires $S$ to be a ``shortest path'' and subpaths of shortest paths are also shortest paths.
As a test procedure for a segment, we use a Dijkstra query.

The optimal path criterion can be generalized to the \emph{personalized path criterion}.
The personalized path criterion requires each trajectory segment to be a personalized path with respect to some driver preference $\alpha$.
This criterion is satisfied if there exists a solution to the LP in~\cref{eq:linear-program}.
Note, that the $\alpha$ for each trajectory segment can differ.

\paragraph{Fixing Edge Cases}
It is possible that there exists a minimal trajectory segment $S \subseteq T$ (consisting of a single road segment) which is not a personalized/optimal path.
One road segment $(u, v)$ might be more expensive in every traversal cost type than another path from $u$ to $v$.
This indicates that the used traversal cost types can not explain driver behavior for taking such an road segment.
For the personalized path criterion, this can be remedied by including a cost type for which each road segment is a personalized path between its source and target intersections.
In general, a unit cost type (every road segment $e$ has $c_{i}(e)=1$) has this property.
This guarantees segmentability for arbitrary trajectories and makes the personalized path more robust.
In our experiments, the number of intersections cost type fulfills this role.

This does, however, not fix the special case of self-loop edges, which in our data set typically represent road segments that allow traversals in parking lots.
Such road segments can never be optimal because the optimal path from the source intersection to itself remains at the intersection.
Self-loop edges can either be dealt with by deleting them from the trajectories, if they do not cover significant areas in the road networks, or by representing such road segments as two edges that each represent partial traversal of the self-looping road segment.

\subsection{Experiments}\label{sec:trajectory-segmentation-experiments}
We now investigate the capabilities of the trajectory segmentation method to identify via-points in trajectories on the basis of the trajectory data set described in \cref{sec:data-set}.
In particular, we use the stitched trajectory set to evaluate our trajectory segmentation approach, \ac{ppts}, to the \ac{opts} baseline which uses only a single cost type to check for the optimal path criterion.
We consider the four variants \ac{opts}-TT, \ac{opts}-Con, \ac{opts}-Int, and \ac{opts}-Cro, that use the travel time, congestion level, number of intersections, and crowdedness, respectively, as the single cost type.
%In addition, we also use the \ac{cts} baseline  which maximizes \ac{brr} (at the expense of \ac{sr}) by placing a segmentation points s.t.\ each trajectory segment consists of exactly one road segment.

We compare \ac{ppts}'s ability to segment trajectories to that of the baselines.
Our comparison is both in terms of the number of trajectories that can be segmented and the ability of the trajectory segmentation algorithms to recover the break points in the stitched trajectory set.
As mentioned in \cref{sec:stitched-trajectories}, these break points indicate a break of $3$ to $33$ minutes and are therefore likely to indicate a via-point within the trajectory.
We discard the self-loop edges within each trajectory to increase segmentability, as described in~\cref{sec:personalized-path-criterion}.
Typically, these self-loop edges represent road segments that allow driving around parking lots.

All algorithms used in our experiments are implemented in the Rust programming language\footnote{\url{https://www.rust-lang.org/}}. We make the implementation of our method, the used graph and some example trajectories publicly available\footnote{\url{https://github.com/Lesstat/ppts}}.
We use \acp{ch}~\cite{FLS17} to speed up the Dijkstra queries by orders of magnitude.
% The trajectory segmentation process uses only the travel time metric and takes $11$ milliseconds per trajectory on average.
% We discuss the relationship between processing time and trajectory length in \cref{sec:processing-time}.

\subsubsection{Evaluation Functions}
We use several evaluation functions to evaluate our trajectory segmentation method and for comparison with the baselines.

\paragraph{Segmentability Score}
The segmentability score, or simply S-score, measures the proportion of trajectories that are segmentable by a trajectory segmentation algorithm.
Ideally, the S-score is $100\%$ indicating that all trajectories in the data set could be segmented by the used trajectory segmentation algorithm.

\paragraph{Break Recovery Rate}
The \ac{brr} is a measure of how good a trajectory segmentation algorithm is at placing segmentation points s.t.\ they coincide with known break points in the stitched trajectories.
Let $\mathit{BP}$ denote the set of known break points in a trajectory $T$ and let $\mathit{SP}$ denote the set of segmentation points output by a trajectory segmentation algorithm that has been given trajectory $T$ as input.
Then, the \ac{brr} of trajectory $T$ is
\begin{equation*}
	\text{BRR}(T)=\frac{|\mathit{RBP}|}{|\mathit{BP}|}
\end{equation*}
where $\mathit{RBP} = \mathit{BP} \cap \mathit{SP}$ is the set of recovered break points.
%A break point $\mathit{bp} \in \mathit{BP}$ is recovered if $\mathit{bp} \in \mathit{SP}$.

\paragraph{Segmentation Rate}
A trajectory segmentation algorithm can achieve a high \ac{brr} by simply segmenting a trajectory into trajectory segments consisting of one road segment each.
Although such a segmentation is guaranteed to recover all break points, it is also very likely to contain a lot of noise in the form of many false positives or false break points.
To measure such noise, we use the \ac{sr} which measures the number of segmentation points per break point:
\begin{equation*}
	\text{SR}(T) = \frac{|\mathit{SP}|}{|\mathit{BP}|}
\end{equation*}
Ideally, the \ac{sr} should be $1$ for a trajectory segmentation that recovers all break points, i.e., has a \ac{brr} of $100\%$.

\paragraph{Segmentation Quality Score}
The segmentation quality score, or simply SQ-score, is a summary score to measure the overall quality of a trajectory segmentation.
It combines the \ac{brr} and \ac{sr} as follows:
\begin{equation*}
	\text{SQ}(T) = \frac{\text{BRR}(T)}{\text{SR}(T)}
\end{equation*}
Note, that the unit of the SQ-score is recovered break points per segmentation point and should ideally be $1$.

\subsubsection{Results}
The results of our experiments are shown in \cref{tab:segmentation-results}.

\paragraph{Segmentability}
As shown in \cref{tab:segmentation-results}, \ac{ppts} and \ac{opts}-Int are both capable of segmenting all trajectories and achieve an S-score of $100\%$.
This result is not too surprising, since both algorithms use a unit cost type---the number of intersections---which guarantees that any trajectory can be segmented by these approaches, as discussed in \cref{sec:personalized-path-criterion}.
The remaining algorithms cannot segment a large portion of the trajectories (more than half in the case of \ac{opts}-Con) and therefore achieve comparatively low S-scores.
Thus, the inclusion of additional cost types can increase segmentability.

\begin{table}
	\centering
	\caption{Mean algorithm performance on all stitched trajectories (ALL) and the $\mathbf{60,249}$ commonly segmentable trajectories (CS) that can be segmented by all algorithms.\label{tab:segmentation-results}}
	\begin{tabular}{lllllll}
		\toprule
		                 & \multicolumn{2}{c}{\emph{ALL}}      &                                    & \multicolumn{3}{c}{\emph{CS}}                                                                                                                  \\
		%& BRR & Segmentability && BRR & Mean No.\ of Segmentation Points \\
		\cline{2-3} \cline{5-7}
		\emph{Algorithm} & \multicolumn{1}{l}{\emph{\ac{brr}}} & \multicolumn{1}{l}{\emph{S-score}} &                               & \multicolumn{1}{l}{\emph{\ac{brr}}} & \multicolumn{1}{l}{\emph{\ac{sr}}} & \multicolumn{1}{l}{\emph{SQ-score}} \\
		\midrule
		\ac{ppts}        & $57.98\%$                           & $\mathbf{100.0}\boldsymbol{\%}$    &                               & $56.29\%$                           & $\mathbf{2.39}$                    & $\mathbf{0.235}$                    \\
		\ac{opts}-TT     & $34.62\%$                           & $58.16\%$                          &                               & $\mathbf{58.35\%}$                  & $2.83$                             & $0.206$                             \\
		\ac{opts}-Con    & $29.32\%$                           & $49.61\%$                          &                               & $57.49\%$                           & $3.99$                             & $0.144$                             \\
		\ac{opts}-Int    & $\mathbf{59.19}\boldsymbol{\%}$     & $\mathbf{100.0}\boldsymbol{\%}$    &                               & $57.61\%$                           & $4.37$                             & $0.132$                             \\
		\ac{opts}-Cro    & $35.86\%$                           & $61.11\%$                          &                               & $58.10\%$                           & $5.62$                             & $0.103$                             \\
		%\ac{cts}      & $100.0\%$  & $100.0\%$  && $100.0\%$  & $67.0$ & $0.015$  \\
		\bottomrule
	\end{tabular}
\end{table}

\paragraph{Segmentation Quality}
As shown in \cref{tab:segmentation-results}, \ac{ppts} and \ac{opts}-Int achieve similar \ac{brr}s that are substantially higher than the remaining \ac{opts} variants.
However, for a fair comparison that ignores the ability of the algorithms to segment trajectories, we have computed \ac{brr}s, \ac{sr}s, and SQ-scores on the subset of trajectories that are commonly segmentable, i.e., the trajectories that can be segmented by all algorithms.
On this subset, the \ac{brr}s of all algorithms are comparable.
This suggests that the superior \ac{brr} when considering all trajectories for \ac{ppts} and \ac{opts}-Int can largely be attributed to their greater capability for segmenting trajectories.

Although the \ac{brr}s are quite similar on the commonly segmentable trajectories, the \ac{sr}s are quite different, as shown in \cref{tab:segmentation-results}
In particular, the \ac{opts}-Con, \ac{opts}-Int, and \ac{opts}-Cro algorithms have, respectively, $67\%$, $83\%$, and $135\%$ more segmentation points per break point than \ac{ppts}.
The \ac{sr} of \ac{opts}-TT algorithm is just $18\%$ higher than that of \ac{ppts}.

The higher \ac{sr}s of the single-cost-type baselines compared to \ac{ppts} suggest that the inclusion of driving preferences and multiple criteria reduces the amount of false positives.
The \ac{ppts} achieves the lowest \ac{brr} on the commonly segmentable trajectories, but, as shown in \cref{tab:segmentation-results}, \ac{ppts} achieves the best overall trajectory segmentation quality with an SQ-score of $0.235$, since it introduces the fewest false break points and thus have the lowest \ac{sr}.
Conversely, \ac{opts}-TT achieves the highest \ac{brr} score on the same data subset, but introduces more false break points than \ac{ppts}.
As a result, \ac{opts}-TT achieves only the second-highest segmentation quality with an SQ-score of $0.206$.
Still, our results support the wide-spread use of the travel time cost type in many routing services, but also show that taking additional cost types and driving preferences into account can lead to better trajectory segmentation.

For the sake of brevity, we present only the comparison between \ac{ppts} and the best-performing baseline, \ac{opts}-TT, in the remainder of this section.

%The quality of the segmentation is largely independent of trajectory length (in number of road segments) and the number of breaks in a trajectory.
%For \ac{fpts}, we found only a weak tendency for the \ac{brr} to lower as trajectory length and the number of breaks in a trajectory increases, with Pearson correlations of $r=-0.24$ and $r=-0.23$, respectively.
%We found no such correlation for \ac{ppts}.

%Although our results are not definite, they suggest that the use of personalized routing and driving preferences allow \ac{ppts} to model driver behavior better than \ac{fpts} resulting in a greater number of recovered break points.

\paragraph{Segmentation Point Accuracy}
\begin{figure}[h]
	\begin{subfigure}{\linewidth}
		\includegraphics[width=\linewidth]{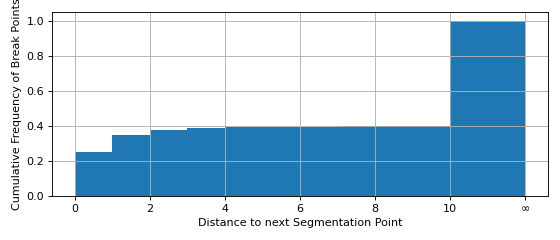}
		\caption{\ac{opts}-TT\label{fig:single-metric-break-decomp-dist}}
	\end{subfigure}
	\begin{subfigure}{\linewidth}
		\includegraphics[width=\linewidth]{%
			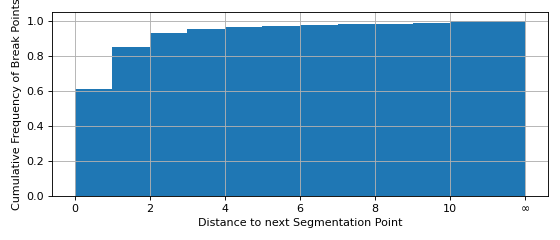}
		\caption{\ac{ppts}\label{fig:multi-metric-break-segment-dist}}
	\end{subfigure}
	\caption{Distribution of distance between a break point and the next segmentation point for (a) \ac{opts}-TT and (b) \ac{ppts}.\@
		Break points in trajectories without any segmentation point are assigned distance $\boldsymbol{\infty}$.%
		\label{fig:break-segment-dist}}
\end{figure}
We have thus far only considered exact break point recovery, but a segmentation may still be useful if it indicates that a break point is near.
\cref{fig:break-segment-dist} shows the percentage of break points which is within a certain (hop) distance to the next segmentation point for \ac{opts}-TT and \ac{ppts}.\@
\ac{ppts} places segmentation points considerably more accurate than \ac{opts}-TT.\@
\ac{ppts} places more than $60\%$ of the break points within one road segment of the nearest segmentation point and over $80\%$ are within two road segments of the next segmentation point.
\ac{opts}-TT achieves less than half of \ac{ppts}'s performance.
However, the performance disparity illustrated in \cref{fig:break-segment-dist} is largely due to better segmentability of trajectories when using \ac{ppts}.
If the analysis is restricted to break points with distance $d < \infty$ to the nearest segmentation point, i.e., trajectories that are segmentable by \ac{opts}-TT, their distributions are comparable.

\paragraph{Qualitative Segmentation Assessment}
A good trajectory segmentation marks break points (or other interesting points) along a trajectory with a segmentation point.
However, a good trajectory segmentation should also avoid too many false positives.

The \ac{ppts} and \ac{opts}-TT, respectively, have an \ac{sr} of $2.39$ and $2.83$ segmentation points per break point, respectively.
However, although these numbers suggest that there are more false break points when using \ac{opts}-TT, our data only contains positive examples of interesting behavior within the trajectory, i.e., the break points in the stitched trajectories.
As result, we cannot quantitatively determine whether the segmentation points that do not match a break point are indeed false positives or mark interesting, but unknown, behavior during the trajectory.
We therefore qualitatively assess the validity of the segmentation of a few trajectories.

\begin{figure}
	\centering
	\begin{subfigure}{\linewidth}
		\centering
		\includegraphics[width=0.95\linewidth]{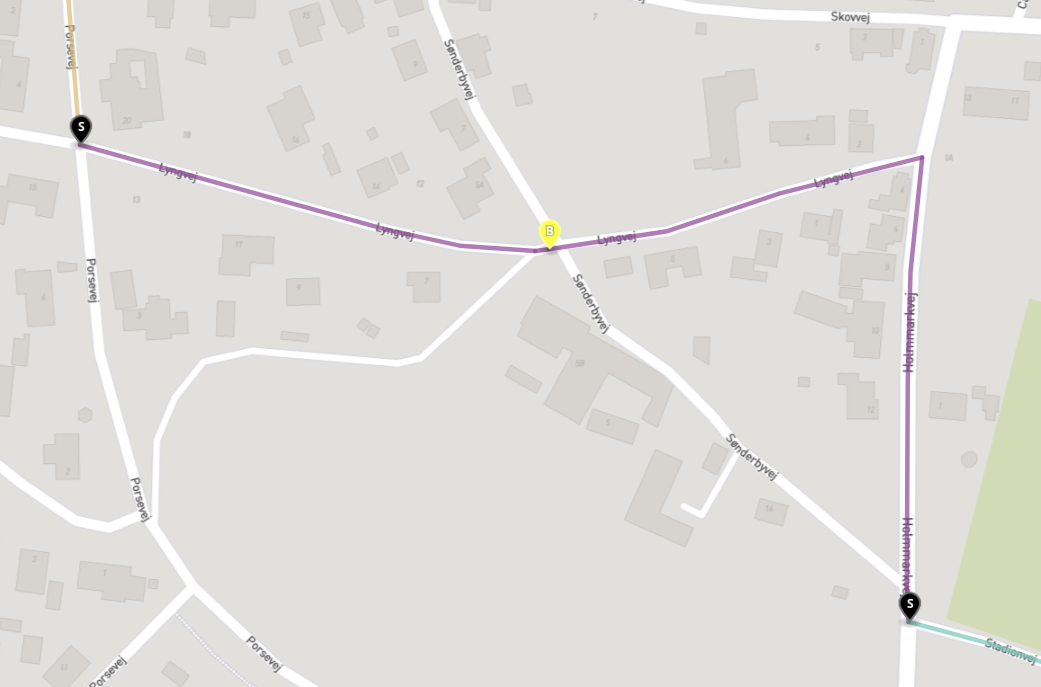}
		\caption{\ac{opts}-TT\label{fig:qualitative-example-1-fpts}}
	\end{subfigure}
	\begin{subfigure}{\linewidth}
		\centering
		\includegraphics[width=0.95\linewidth]{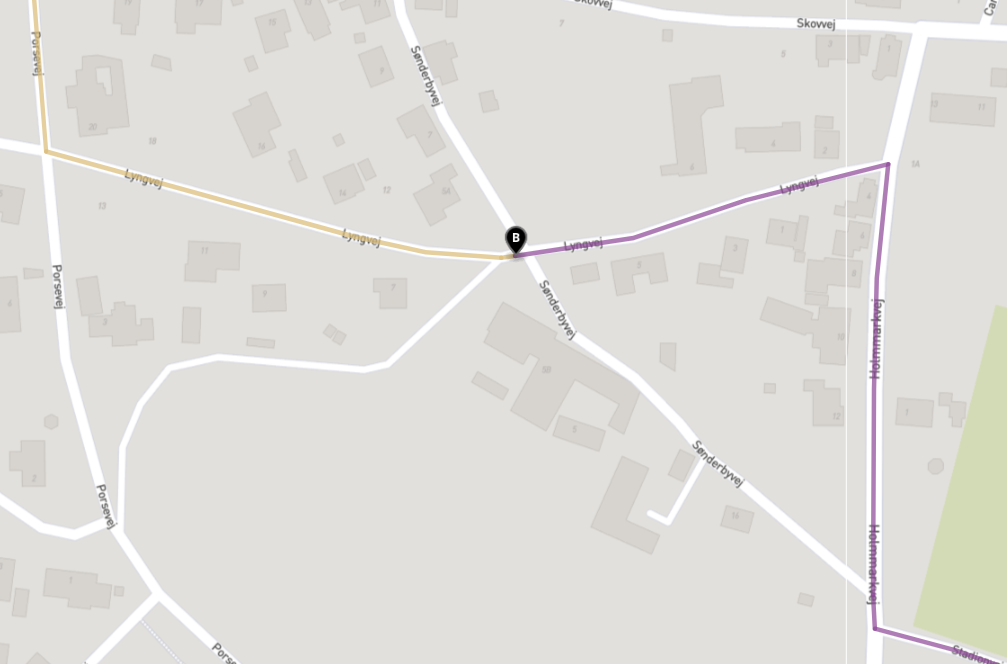}
		\caption{\ac{ppts}\label{fig:qualitative-example-1-ppts}}
	\end{subfigure}
	\caption{%
		A break point in a trajectory and the segmentation points for (a) \ac{opts}-TT and (b) \ac{ppts}.
		Yellow markers labeled `B' indicate a break point and black markers labeled `S' a segmentation point.
		A black marker labeled B indicates a break point that is recovered by a segmentation point.\label{fig:qualitative-example-1}}
\end{figure}

\cref{fig:qualitative-example-1} shows a break point (marked with `B' in yellow) in a segmented trajectory.
The segmentation by \ac{opts}-TT shown in \cref{fig:qualitative-example-1-fpts} places two segmentation points (marked with `S' in black) around the break point.
These segmentation points fail to recover the break point but are both within a distance of two road segments of the break point.
Thus, the \ac{opts}-TT segmentation appear to detect the presence of the break point, but fails to place the segmentation points exactly.
The \ac{ppts} segmentation shown in \cref{fig:qualitative-example-1-ppts}, is a better segmentation and recovers the break point exactly (indicated by the black marker labeled `B').

\begin{figure}
	\includegraphics[width=0.95\linewidth]{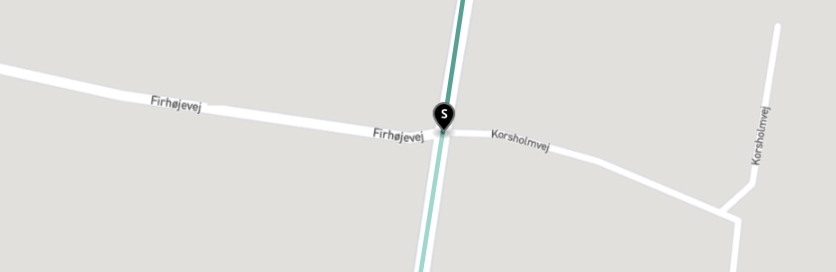}
	\caption{A segmentation point with no obvious event occurring.\label{fig:qualitative-example-2}}
\end{figure}

\cref{fig:qualitative-example-2} shows another part of the trajectory shown in \cref{fig:qualitative-example-1}.
Here, \ac{opts}-TT places a segmentation point without comparable segmentation points in the \ac{ppts} segmentation. This additional segmentation point has no apparent meaning, and, upon detailed inspection, appears to occur due to inaccuracies in the estimated travel time in the area.
This suggests that \ac{ppts} may be more robust than \ac{opts}-TT to noise in the traversal cost data.

\begin{figure}
	\includegraphics[width=0.95\linewidth]{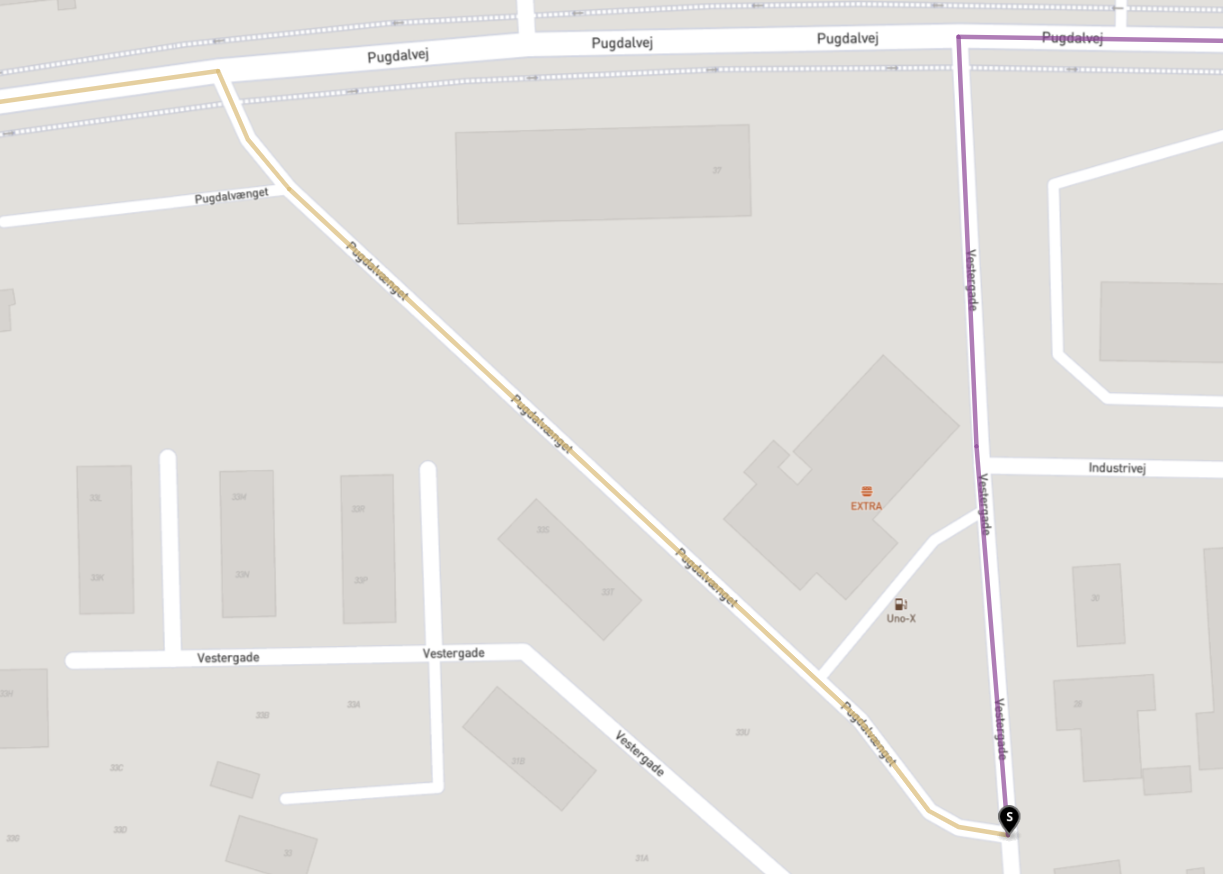}
	\caption{A segmentation point recovers a detour to a gas station that is not marked as a break in our data set.\label{fig:single-metric-gas-station}}
\end{figure}

For the purposes of quantitative evaluation, our method attempts to recover breaks of $3$ to $33$ minutes from trajectories.
However, our trajectory segmentation approach can discover interesting behavior beyond these known breaks.
For instance, \cref{fig:single-metric-gas-station} shows a segmentation point marking a detour to a gas station.
This segmentation point is placed by both \ac{opts}-TT and \ac{ppts}.

\subsubsection{Processing Time}\label{sec:processing-time}
% \begin{figure}
%   \begin{subfigure}{\columnwidth}
%    \includegraphics[width=\linewidth]{%
%      figures/travel-time-metric-segmentation-run-time.png}
%    \caption{\ac{fpts}}
%   \end{subfigure}
%   \begin{subfigure}{\columnwidth}
%    \includegraphics[width=\linewidth]{%
%      figures/four-metrics-segmentation-run-time.png}
%    \caption{\ac{ppts}}
%   \end{subfigure}
%   \caption{Processing time for (a) \ac{fpts} and (b) \ac{ppts}.\label{fig:run-time-comparison}}
% \end{figure}

While using personalized routing does improve break recovery, it comes with an increase in processing time of trajectory segmentation.
The increase in processing time for the personalized path variant is mostly driven by the CH-Dijkstra queries being slower.

The trajectory segmentation process is trivially parallelizable, since each trajectory can be processed independently, making segmentation of even billions of trajectories feasible.
In our experiments, we parallelized the trajectory segmentation process across $64$ cores, each with a clock speed of $2.3$ GHz.
The time to process the $190,199$ stitched trajectories for single-criteria and multi-criteria trajectory segmentation is, respectively, $1$ and $5$ hours in total, and $19$ and $95$ milliseconds per trajectory on average.
The total processing time took about half an hour in wall-clock time.

% We assume that the severe outliers with up to $13s$ processing time are artifacts of the parallelization.
% When we investigated some of these outliers in a separate run the processing time was closer to the mean processing time.

\subsection{Discussion}
Overall, \ac{ppts} achieves the highest trajectory segmentation quality in our experiments, followed by \ac{opts}-TT.\@
The results suggest that \ac{ppts} has two primary advantages over the baselines in our experiments.
The use of multiple cost types and driving preferences makes \ac{ppts} capable of (1) segmenting more trajectories and (2) explaining driving behavior better, resulting in fewer false break points being placed.
Our qualitative assessment of the \ac{opts}-TT and \ac{ppts} segmentations support the conclusion that the segmentation points placed by \ac{ppts} are less likely to be false positives.
In addition, \ac{ppts} discovered a detour to a gas station that is not indicated by a break point in our trajectory data set.

Even in the case where a break point is not recovered, \ac{ppts} is likely to place a segmentation point near the break point.
\ac{ppts} places a segmentation point within a distance of $3$ road segments for $95\%$ of break points, but \ac{opts}-TT for less than $40\%$ of the break points.
Although, this difference is largely explained by \ac{ppts} being capable of segmenting more trajectories, it suggests that \ac{ppts}'s segmentation points are likely to indicate some interesting part of a trajectory.
Although the increase in performance of \ac{ppts} over \ac{opts}-TT comes at a factor $5$ increase in processing time, it is still capable of segmenting a trajectory in a fraction of a second on average.

\paragraph{Stitching Parameters}
Changing the parameters for the stitching process has only very little effect on our results and do not affect our conclusions.
The results are virtually invariant to changes to the temporal stitching threshold.
However, they are sensitive to changes to the stitch length threshold, although the effect is minor.
The longer the stitches are allowed to be, the worse break recovery performance for both \ac{opts}-TT and \ac{ppts}.
This is likely because more noise is introduced when longer stitches are allowed.

\section{Robust Driving Preference Mining}%
\label{sec:robust-driving-preference-recovery}
The trajectory segmentation approach presented in \cref{sec:trajectory-segmentation} implicitly recovers driving preferences $\alpha$ for each trajectory segment when determining the personalized path segments by searching for a solution to \cref{eq:linear-program}.
%Our trajectory segmentation relies on \cref{eq:linear-program} to be brittle s.t.\ segmentation points can be placed when a driver deviates from the optimal route.
Although it is possible to collect the preference vectors for each trajectory segment, one is typically more interested in a \emph{single} preference vector to describe a driver's general behavior for use in, e.g., personalized route planning~\cite{FS15,Balteanu13,YangPersonalizedRouting}.
Fortunately, this is possible with only a minor modification of the linear program in \cref{eq:linear-program}:
\begin{equation}\label{eq:linear-program-driving-preference-recovery}
	\begin{aligned}
		\textsc{Minimize} \qquad \ \ \, & \delta                                                                   \\
		\textsc{subject to} \qquad      & \forall \pi \in \Pi \colon  p(T \mid \alpha)-p(\pi \mid a)   \leq \delta \\
		                                & \sum_{i=1}^d \alpha_i =1                                                 \\
		                                & \forall i \in \{1, \dots, d\}\colon \alpha_i \geq 0                      \\
		                                & \delta \geq 0
	\end{aligned}
\end{equation}
As with \cref{eq:linear-program}, this linear program may also be solved in polynomial time using the LP path oracle described in \cref{sec:lp-path-oracle}.

The following modifications have been made to \cref{eq:linear-program-driving-preference-recovery}.
First, the solution to the linear program is a preference vector $\alpha$ for the whole trajectory $T$ rather than a trajectory segment.
Second, by introducing $\delta$ to the first constraint, $T$ is not required to be a personalized path w.r.t.\ to the $\alpha$.
Third, \cref{eq:linear-program} minimizes $\delta$ s.t.\ $T$ is as close to being shortest-path optimal as possible w.r.t.\ to the preference vector $\alpha$ that is the solution to the linear program.

The effect of the modifications made in \cref{eq:linear-program-driving-preference-recovery} is that the linear program always has a feasible solution and therefore our approach always outputs some preference vector $\alpha$.
If $\delta = 0$, then the recovered $\alpha$ fully explains the driver behavior in the trajectory and is identical to the solution of \cref{eq:linear-program}.
Otherwise, if $\delta > 0$, then the recovered $\alpha$ does not fully explain the driver behavior but explains it as much as possible given the available traversal costs.

%\begin{figure}
%	\includegraphics[width=0.95\linewidth]{figures/robust-preferences/1073191-repr-path.png}
%    \caption{Illustration of the shortest path (blue) `close' to the trajectory (black) via recovered preference. They exactly overlap except for the non-optimal loop.\label{fig:robust-example}}
%\end{figure}
%Unlike \cref{eq:linear-program}, \cref{eq:linear-program-driving-preference-recovery} still has solutions if deviations from the shortest path w.r.t.\ $\alpha$ are found.
%In essence, the deviations are ignored.
%\cref{fig:robust-example} shows an example of such an ignored deviation.
%The figure shows the end of a trajectory $T=v_0v_1\dots v_{k-1}$ in black.
%The trajectory is overlaid with the \emph{recovered shortest-path} from $v_0$ to $v_{k-1}$.
%This recovered shortest path is computed using the personalized costs w.r.t\ to the $\alpha$ found by solving \cref{eq:linear-program-driving-preference-recovery} for trajectory $T$.
%The blue marker labeled `T' marks the end of the trajectory.
%As shown in the figure, the loop (in black) at the end of the trajectory is ignored in the recovered shortest path.

\subsection{Experiments}\label{sec:acmp-results}
We now evaluate our driving preference mining approach on a personalized routing task using the data set described in \cref{sec:data-set}.
Specifically, we evaluate our approach for each trajectory $T=v_0v_1\dots v_{k-1}$ in a trajectory set as follows.
First, we solve \cref{eq:linear-program-driving-preference-recovery} for $T$.
Then, we use the resulting preference vector $\alpha$ to compute a personalized route (or personalized path) $\pi$ from $v_0$ to $v_{k-1}$.
Ideally, the preference vector $\alpha$ combined with the source $v_0$ and target $v_{k-1}$ is sufficient to reconstruct or recover the route driven in trajectory $T$.
We therefore refer to $\pi$ as the \emph{recovered route} of trajectory $T$.

W.r.t.\ the task of  personalized routing, we are interested in measuring two qualities about our approach.
First, how well do the recovered preference vectors model driving behavior.
Second, how well do the preference vectors match the preferences of the drivers.

\subsubsection{Evaluation Functions}
To measure our approach's ability to model driving behavior, we use the \ac{rrro}:
\begin{equation*}
	\text{RRRO}(T, \pi) = \frac{|T \cap \pi|}{|T|}
\end{equation*}
Let $\alpha$ be a preference vector recovered from trajectory $T$.
Here, $\pi$ is the recovered route of trajectory $T$.
If the preference vector $\alpha$ used to construct $\pi$ fully captures the driving preferences exhibited in $T$, then the route $\pi$ recovered using $\alpha$ should be identical to $T$, resulting in a relative recovered route overlap of $1$.

To measure whether the preference vectors found using our approach match (actual) driver preferences, we use the \ac{rcrs}.
The \ac{rcrs} reflects the view that two routes are equivalent if their personalized costs are identical:
\begin{equation*}
	\text{RCRS}(T, \pi) = \frac{p(\pi \mid \alpha)}{p(T \mid \alpha)}
\end{equation*}
Note, that $p(T \mid \alpha) \geq p(\pi \mid \alpha)$ since $\pi$ is the shortest path w.r.t.\ the personalized costs of the road segments for the given preference vector $\alpha$.
Thus, the \ac{rcrs} is always between $0$ and $1$.
An \ac{rcrs} value of, e.g., $0.8$, indicates that the preference vector $\alpha$ accounts for $80\%$ of the personalized cost of trajectory $T$.
If the preference vector $\alpha$ fully captures the driver's preferences, then the \ac{rcrs} is $1$.

\subsubsection{Baselines}
We refer to our approach as \ac{rdp} and compare its performance with two baselines.
The first baseline, \ac{ttp}, always returns a preference vector that has weight one for travel time.
The second baseline, \ac{brp}, generates five random preference vectors for a trajectory, evaluates them and returns the preference with the best result.
The \ac{brp} baseline is run independently for the two evaluation functions used in our experiments.

\subsubsection{Results}
We run our experiment on both the unstitched and stitched trajectory sets.
The results on both trajectory sets are similar.
For brevity, we report only the results on the unstitched trajectory set.

We summarize the results on the unstitched trajectories in \cref{tab:acmp-results}.
As shown in the table, \ac{rdp} achieves both the highest mean \ac{rrro} and mean \ac{rcrs} that are, respectively, $5.7\%$ and $2.4\%$ better than the best performing baseline $\ac{ttp}$. We expect that these figures will be even higher if more than four traversal cost types are used.

\begin{table}
	\centering
	\caption{Mean \ac{rrro} and \ac{rcrs} of \ac{rdp}, \ac{ttp}, and \ac{brp} of different algorithms for personalised routing on the unstitched trajectory set.\label{tab:acmp-results}}
	\begin{tabular}{lccc}
		\toprule
		          & \emph{\ac{rdp}} & \emph{\ac{ttp}} & \emph{\ac{brp}} \\
		\midrule
		\ac{rrro} & $\mathbf{0.74}$ & $0.70$          & $0.66$          \\
		\ac{rcrs} & $\mathbf{0.87}$ & $0.85$          & $0.81$          \\
		\bottomrule
	\end{tabular}
\end{table}

\begin{figure}
	\begin{subfigure}{\columnwidth}
		\includegraphics[width=\linewidth]{%
			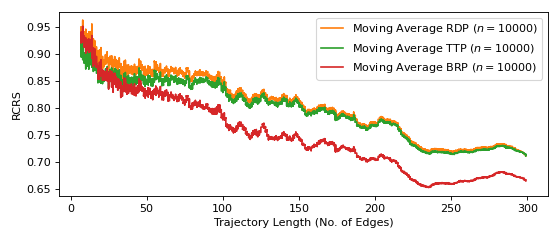}
		\caption{\ac{rcrs}}
	\end{subfigure}
	\begin{subfigure}{\columnwidth}
		\includegraphics[width=\linewidth]{%
			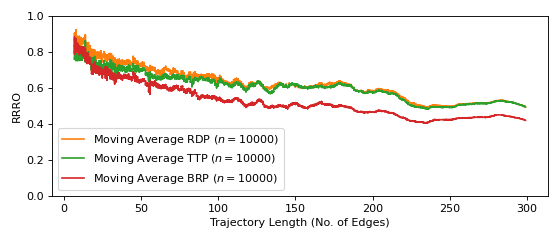}
		\caption{\ac{rrro}}
	\end{subfigure}
	%\begin{subfigure}{\columnwidth}
	%	\includegraphics[width=\linewidth]{%
	%		figures/robust-preferences/unstitched-wrong-turns-moving-average.png}
	%	\caption{No.\ of Wrong Turns}
	%\end{subfigure}
	\caption{The (a) \ac{rcrs} and (b) \ac{rrro} scores of \ac{rdp}, the \ac{ttp} and the \ac{brp} with unstitched trajectories.\label{fig:acmp-results}}
\end{figure}

The \ac{rdp} shows superior performance compared to both baseline algorithms, but the performance of both \ac{rdp} and the baselines deteriorate as trajectory length increases as shown in \cref{fig:acmp-results}.
This is not particularly surprising since longer trajectories are more likely to contain more via-points, and none of the approaches in this experiment take such information into account.
In addition, \ac{ttp} approaches the performance of \ac{rdp} as trajectory length increases.
This suggests that people are likely to prioritize travel time more on long trips which matches both our expectations and anecdotal experiences.

% TODO: May be worth having this as a separate section using the figure currently in the method description.
%\subsubsection{Robustness to Deviations}

%As a result of the linear program formulation in our approach, tend to be ignored to some extent.deviations in we discussed in \cref{}
%As mentioned in \cref{sec:robust-driving-preference-recovery}, there may exist routes that are the preference vector $\alpha$ is not required to yield 

%\subsubsection{Stitched Trajectories}
%The results of the experiments with stitched trajectories are worse than with unstitched trajectories. For instance, the relative edge overlap of the ACMP solution is $0.32$ instead of $0.74$. However, the ACMP solution is still better than the benchmark functions. These results meet our expectations as stitched trajectories are mostly not similar to any optimal path.

\subsubsection{Robustness}
We call our algorithm robust because, in contrast to the original algorithm given in~\cite{FLS16}, it does not require the trajectory to be a personalized path. Therefore, we are able to recover a driving preference for each of the $1,306,392$ trajectories. On the contrary, the baseline algorithm is only able to process $590,251$ trajectories, which is less than half. Hence, our modification is indeed a considerable robustness improvement.

\subsubsection{Processing Time}
% The statistics regarding processing time of the unstitched trajectories are shown in \cref{fig:acmp-processing-time}.
We computed the results with a single core with a clock speed of $2.3$ GHz.
The average processing time is $1.04$ milliseconds per trajectory and $0.02$ milliseconds per road segment, but is proportional to the number of road segments in the trajectory.
%The average processing time increases with the number of road segments in the trajectory.
% The bend at around 600 road segments is likely caused by round trips, i.e., trajectories that have the same source and destination. Round trips may have many road segments. The processing time, however, mainly depends on the length of the optimal paths.
% \begin{figure}[h]
% 	\includegraphics[width=\linewidth]{%
% 		figures/robust-preferences/unstitched-processing-time.png}
% 	\caption{Processing time of the ACMP solver.\label{fig:acmp-processing-time}}
% \end{figure}

\subsection{Discussion}
Our experiments show that the \ac{rdp} preference vectors explain driver behavior better than the \ac{ttp} and \ac{brp} baselines.
Although the average results of \ac{ttp} are very similar to those of \ac{rdp}, they are never better.
This is a strong indication that our approach indeed finds the best preferences to describe the drivers' behavior.
In addition, the fast processing time (and trivial parallelizability) of \ac{rdp} makes it scalable to even very large trajectory data sets.

Notably, the performance gap between \ac{rdp} and \ac{ttp} nearly disappears for long trajectories.
This matches our expectation that travel time is the most important criterion for such trajectories.

\section{Conclusion}\label{sec:conclusion}
In this paper, we have presented two techniques for large scale trajectory segmentation and driver preference mining.
We have shown experimentally that our proposed trajectory segmentation approach is a useful tool for understanding the semantics of a trajectory, e.g., the driver's intentions or changing destinations.
In addition, our experiments showed that our proposed driver preference mining technique can indeed discover driver preferences for real trajectories and is sufficiently robust to process such data.
Our techniques can be implemented efficiently in practice and are trivially parallelizable.
Thus, they scale to very large trajectory sets consisting of millions or even billions of trajectories.

Interestingly, our approaches for trajectory segmentation and driving preference mining rely on the same model of driving preferences, showing that these two tasks are closely linked.
To the best of our knowledge, we are the first to show this link.

\paragraph{Future Directions}
There are many interesting directions for both our trajectory segmentation and driver preference mining approaches.

Our trajectory segmentation approach relies on linear combinations of costs.
However, relationships between costs may be more complex and present an interesting opportunity for future work.
In addition, driver preferences can be highly context-dependent and depend on, e.g., the time of day~\citep{YangPersonalizedRouting}.
Extending our approach to utilize such contextual information is an important future direction.

In our driver preference mining approach, the linear program may have multiple solutions corresponding to a large set of preferences or a preference space.
This makes it difficult to compare recovered preferences among trajectories or trajectory segments of the same trajectory.
In particular, solving the linear program for two trajectories generated by two drivers with the same preferences may yield two different solutions, even if they also follow the same route.
An important future direction is therefore to extend our driving preference mining technique to output identical (or at least similar) preferences in such situations.
This will enable analysis of driver behavior through, e.g., driver preference clustering.

Finally, our driving preference mining approach assumes that the intent of the driver is to go straight from the start of the trajectory to the end of the trajectory.
Thus, it ignores that such trajectories may have via-points.
However, as demonstrated by our experiments, our trajectory segmentation approach can discover such via-points in trajectories.
In future work, it would be interesting to explore synergies between our trajectory segmentation and our driving preference mining approach.
%\COMMENT{Tobias}{Perhaps discussing some of the issues of choosing alphas somewhat arbitrarily from the alpha space should be discussed here in relation to the applications we are considering.}
%One interesting research question whether it is possible to efficiently determine the optimal $\alpha$ such that a given trajectory is decomposed into the fewest optimal segments according to that $\alpha$. We currently determine this $\alpha$ heuristically, and even though it is reasonably close to the optimum (via comparison to the lower bound for the minimum size of a decomposition), this is of algorithmic interest. 

%\ignore{
\section*{Acknowledgments}
This work was supported in part by the DiCyPS project, by grants from the Obel Family Foundation and VILLUM FONDEN, and in part by the Deutsche Forschungsgemeinschaft (DFG) within the priority program 1894: Volunteered Geographic Information.
%}

\bibliographystyle{ACM-Reference-Format}
\bibliography{TrajectoryAnalysis.bib}

%\clearpage 
\section*{Appendix}
\begin{appendices}
	\crefalias{section}{appendix}
	\crefalias{subsection}{appendix}
	\crefalias{subsubsection}{appendix}
	\newcommand{\Tstitched}{\mathbb{T}_{\mathit{stitched}}}
\newcommand{\Tcurrent}{T_{\mathit{current}}}
\newcommand{\taucurrent}{e_{\mathit{current}}}
\newcommand{\GetEndTime}[1]{\textsc{GetEndTime}(#1)}
\newcommand{\ALGINDENT}{\hspace{\algorithmicindent}}

\section{Trajectory Stitching}\label{sec:trajectory-stitching}
Our trajectory dataset $D = \{ \mathbb{T}_1, \dots, \mathbb{T}_n \}$ consists of sets of trajectory sequences of the form $\mathbb{T}_i = (T_1, \dots, T_m)$.
Each trajectory sequence $T_i$ contain trips specific to driver $d_i$ and are in temporal order s.t.\ trajectory $T_j \in \mathbb{T}_i$ started before trajectory $T_{j+1} \in \mathbb{T}_i$.
We define a stitched trajectory data set based on the data set $D$ as
%\begin{equation*}
  $D_{\mathit{stitched}} = \bigcup_{i=1}^n \textsc{StitchTrajectories}(\mathbb{T}_i).$
%\end{equation*}

\begin{algorithm}
  \caption{Trajectory Stitching \label{alg:trajectory-stitching}}

  \begin{algorithmic}[1]
  \Function{StitchTrajectories}{$\mathbb{T} = (T_1, \dots, T_n)$}
    \State $\Tstitched \gets \emptyset$
    \State $\Tcurrent \gets T_1$ \label{line:ts-Tcurrent-init}
    \State $\taucurrent \gets \GetEndTime{\Tcurrent}$ \label{line:ts-taucurrent-init}

    \For{$i=2$ to $n$}\label{line:ts-loop-start}
      \State $s_i \gets \textsc{GetStartTime}(T_{i})$ \label{line:ts-get-start-time}
      \If{$s_i - \taucurrent \leq 30\ \text{minutes}$ and  \label{line:ts-if-start} \\
          \ALGINDENT{} \ALGINDENT{} \ALGINDENT{} $T_i$ and $\Tcurrent$ are pseudo-connected} \label{line:ts-if-cond-end}
          \State{$\Tcurrent \gets \textsc{Stitch}(\Tcurrent, T_i)$} \label{line:ts-stitch}
      \Else
          \State $\Tstitched \gets \Tstitched \cup \{\Tcurrent\}$ \label{line:ts-Tstitched-add}
          \State $\Tcurrent \gets T_i$ \label{line:ts-Tcurrent-update}
      \EndIf
      \State $\taucurrent \gets \GetEndTime{\Tcurrent}$ \label{line:ts-endtime-update}
    \EndFor\label{line:ts-loop-end}
    \State $\Tstitched \gets \Tstitched \cup \{\Tcurrent\}$ \label{line:ts-add-before-return}
    \State \Return $\Tstitched$ \label{line:ts-return}
  \EndFunction
\end{algorithmic}
\end{algorithm}
The $\textsc{StitchTrajectories}$ function, defined in \cref{alg:trajectory-stitching}, takes as input the trajectories of a driver in temporal order.
We use $\Tcurrent$ to keep track of the current trajectory considered for stitching.
Initially, $\Tcurrent$ is set to $T_1$.
We use $\taucurrent$ to keep track of the end time of the current trajectory $\Tcurrent$, i.e., its last recorded GPS point.
In a loop, we scan the input trajectories $\mathbb{T}$ sequentially for stitching candidates, starting from trajectory $T_2$.
We first store the time of the first GPS point associated with trajectory $T_i$ in $s_i$.

We then check whether $T_i$ is both temporally and spatially near enough to stitch with $\Tcurrent$.
The two trajectories $\Tcurrent$ and $T_i$ are temporally near enough to stitch if there is at most a $30$ minute difference between $\taucurrent$ and $s_i$.
Two trajectories $T_1 = (e_1, \dots, e_i)$ and $T_2 = (e_{i+1}, \dots, e_{i+j})$, where $e_i = (u, v)$ and $e_{i+1} = (w, x)$, are pseudo-connected if the shortest route between $v$ and $w$ consists of at most one road segment or is less than $200$ meters in length.
If both stitching conditions are met, $T_i$ is stitched to $\Tcurrent$ by invoking the $\textsc{Stitch}$ function.

For pseudo-connected trajectories $T_1$ and $T_2$, $\textsc{Stitch}$ is defined as
%\begin{equation*}
$
  \textsc{Stitch}(T_1, T_2) = (e_1, \dots, e_i, e'_1, \dots, e'_k, \dots, e_{i+j})
$
%\end{equation*}
where $(e'_1, \dots, e'_k)$ is the shortest route connecting $v$ and $w$ which we refer to as a \emph{stitch}.
Then, the stitched trajectory is assigned to $\Tcurrent$.

If the stitching conditions are not met, we cannot stitch more trajectories to $\Tcurrent$.
We then add the current trajectory to $\Tstitched$ and let $T_i$ be the new current trajectory.
% The end time of the current trajectory $e_i$ is updated before the next iteration of the loop.
Note that after the first iteration $\Tcurrent$ may be a stitched trajectory.
For a stitched trajectory $T' = \textsc{Stitch}(T_1, T_2)$, we define $\textsc{GetEndTime}(T') = \textsc{GetEndTime}(T_2)$.
After scanning through all of the input trajectories, we add the last trajectory to $\Tstitched$ and finally return the stitched trajectories.

	\section{Routing Cost Type Details}
In this section, we describe how the travel time, congestion, and crowdedness routing costs are derived in further details.

\subsection{Travel Time}\label{sec:travel-time-derivation}
The vehicle trajectories in our trajectory set have the tendency to be concentrated on a few popular segments, as such, many road segments have few or no traversals in the trajectory set.
We therefore require a means of estimating travel times for such road segments.
To this end, we use a pre-trained machine learning model to provide travel time estimates.
However, for road segments with an abundance of traversal data the model's estimates may be inaccurate.
Inspired by previous work~\cite{estimation-updates}, we therefore combine travel time estimates with travel times of historical traversals s.t.\ when the driving speed estimate of a road segment becomes increasingly less influential the more historical traversals the road segment is associated with.

We compute the travel time $t_e$ for a road segment $e$ as
%\begin{equation*}
  $t_e = \frac{k\hat{t_e} + n\bar{t_e}}
              {k+n}$
%\end{equation*}
where $\hat{t_e}$ is the estimate of the mean travel time, $\bar{t_e}$ is the mean travel time of the historical traversals, $n$ is the number of historical traversals of segment $e$ in the trajectory dataset, and $k$ represents the confidence in $\hat{t_e}$.
We use $k=10$ in our experiments.

We use a pre-trained \ac{rfn}~\cite{rfn-long} to provide travel time estimates $\hat{t_e}$ for each road segment $e \in E$.
Specifically, we use the best performing \ac{rfn} from~\cite{rfn-long} which has been trained on the Danish Municipality of Aalborg using trajectories within the municipality that occurred between January 1st 2012 and June 30th 2013. Despite having been trained only on a subset of the network, the model generalizes well to unseen areas of the road network~\cite{rfn-long}. However, in a few cases the network would give very low values. We therefore modify the output s.t.\ the estimated driving speed on any road segment cannot be below $5$ kmh.

\subsection{Congestion}\label{sec:congestion-derivation}
We assign a congestion level to road segment $e$ depending on the speed limit $s_e$ on the segment in km/h, the length of $l_e$ of the segment in km, and the travel time $t_e$ in hours.
Let $\tau_e = l_e / s_e$ denote the travel time on road segment $e$ if a vehicle is driving at exactly the speed limit.
Formally, we assign road segment $e$ the congestion level
%\begin{equation*}
  $c_e = \max\{1 - \frac{t_e}{\tau_e}, 0\}$
%\end{equation*}
s.t.\ a value of $0$ indicates that it is possible to drive at (or above) the speed limit and a value of $1$ indicates that the road segment is not traversable.

The value of $\tau_e$ relies on the speed limit of road segment $e$.
We use a speed limit data set that combines \ac{osm} speed limits with speed limits provided by Aalborg Municipality and Copenhagen Municipality~\cite{rne-workshop}.
This data set contains $163\,044$ speed limits, thus leaving many road segments without a known speed limit.
In such cases, we use an \ac{osm} routing heuristic\footnote{See \url{https://wiki.openstreetmap.org/wiki/OSM_tags_for_routing/Maxspeed}.} which in Denmark assigns a speed limit of $130$ km/h to motorways, a speed limit of $50$ km/h in cities, and a speed limit of $80$ km/h on other types of segments.
For our data, we count a road segment as in a city if either the source or destination intersection is in a city according to its attributes.

\subsection{Crowdedness}\label{sec:crowdedness-derivation}
This routing cost type describes how `crowded' the landscape around a road segment is.
It is derived from the number of \ac{osm} nodes in the vicinity of the road segment.
We use all \ac{osm} nodes in Denmark from a 2019 data set regardless wether they represent a road, a building or some other point of interest.
To calculate it, we first overlay our graph with a grid and count the \ac{osm} nodes within each cell.
For each road segment, we locate the \ac{osm} nodes that are part of its geometry in the grid.
The cost per road segment is then the sum of the cell counts of its (geometry) nodes.
We use a grid of $2000$ by $2000$ resulting in a cell size of roughly 209m x 177m.

\end{appendices}

\end{document}